\documentclass{article}
\usepackage{iclr2026_conference,times}
\usepackage[hyphens]{url}
\usepackage{graphicx}
\usepackage{flafter}
\usepackage{amsmath}
\usepackage{amssymb}
\usepackage{caption}
\usepackage{booktabs}
\usepackage{algorithm}
\usepackage{algorithmic}
\usepackage{placeins}
\usepackage{xcolor}
\usepackage{hyperref}
\definecolor{hunyuanblue}{HTML}{0052D9}
\hypersetup{
    breaklinks=true,
    citecolor=hunyuanblue,
    colorlinks=true,
    linkcolor=hunyuanblue,
    urlcolor=hunyuanblue
}
\fancypagestyle{bodypages}{
    \fancyhf{}
    \fancyfoot[C]{\thepage}
    \renewcommand{\headrulewidth}{0pt}
    
}
\fancypagestyle{firstpage}{
    \fancyhf{}
    \lhead{\includegraphics[height=20pt]{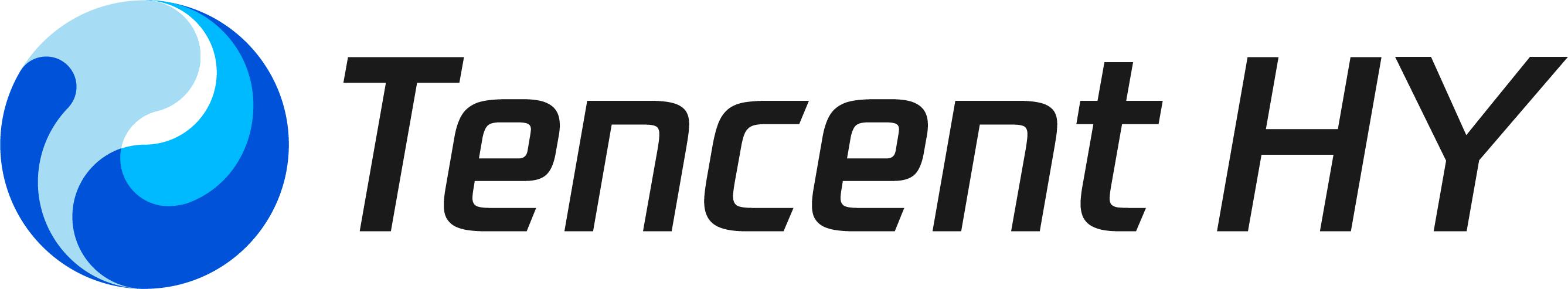}}
    \rhead{\textcolor[HTML]{4D4D4D}{\today}}
    \renewcommand{\headrulewidth}{1pt}
}
\makeatletter
\def\@maketitle{%
\vbox{\hsize\textwidth
  \centering
  {\fontsize{17}{20}\selectfont\bfseries \@title\par}
  \vskip 1em
  \ificlrfinal
    \lhead{}%
    {\normalsize \@author\par}
  \else
    \lhead{Under review as a conference paper at ICLR 2026}%
    {\normalsize Anonymous authors\\Paper under double-blind review\par}
  \fi
  \vskip 0.22in minus 0.08in}}
\makeatother

\title{Deep Research Pretraining via Predictive Navigation}
\author{%
\begin{tabular}{@{}l@{\hspace{1.5em}}l@{\hspace{1.5em}}l@{\hspace{1.5em}}l@{\hspace{1.5em}}l@{\hspace{1.5em}}l@{}}
Jiang Zhou &
Zhiyuan Fan &
Xing Wu\thanks{Corresponding Authors. Correspondence to \texttt{\{ucaswu,maxwellyu\}@tencent.com}.} &
Tinghao Yu\footnotemark[2] &
Feng Zhang &
Lilin Wang \\
\multicolumn{6}{@{}c@{}}{\textbf{Hunyuan Team, Tencent}}
\end{tabular}
}

\iclrfinalcopy
\begin{document}

\setcounter{footnote}{1}
\maketitle
\pagestyle{bodypages}

\begin{abstract}
Deep research agents are often trained on expensive, environment-grounded tool-use trajectories that require repeated retrieval, document inspection, and report evaluation. We introduce Deep Research Pretraining (DRP), an offline framework that derives predictive-navigation supervision from naturally occurring evidence structures. Given a citation-bearing or hyperlinked passage, DRP constructs a proxy research objective, recovers linked evidence and graph-related alternatives, and converts them into search--open--write trajectories. This teaches models what to search for, which documents to inspect, and how to synthesize evidence, without a live retrieval environment or executed policy rollout. We instantiate DRP on scholarly citation graphs (DRP-Paper) and Wikipedia hyperlinks (DRP-Web), continually pretrain separate Qwen3-14B-Base models on 1B tokens, and fine-tune them on controlled fractions of 13K agent trajectories. Across five independently sampled subsets at each low-data budget, both variants consistently outperform matched no-DRP models on DeepResearch Bench. With one quarter of the SFT data, DRP-Web even surpasses a fixed no-DRP full-data checkpoint, with gains transferring to ResearchQA, WebWalkerQA, and SimpleQA. Starting from matched low-data SFT checkpoints, the DRP-Web advantage also persists through subsequent agentic RL. Source-matched and evidence-mismatch controls indicate that these improvements arise from evidence-conditioned navigation rather than domain exposure or agent-format imitation. DRP thus provides a promising complementary approach to trajectory-based agent training.
\end{abstract}

\section{Introduction}

\begin{figure*}[t]
\centering
\includegraphics[width=\textwidth]{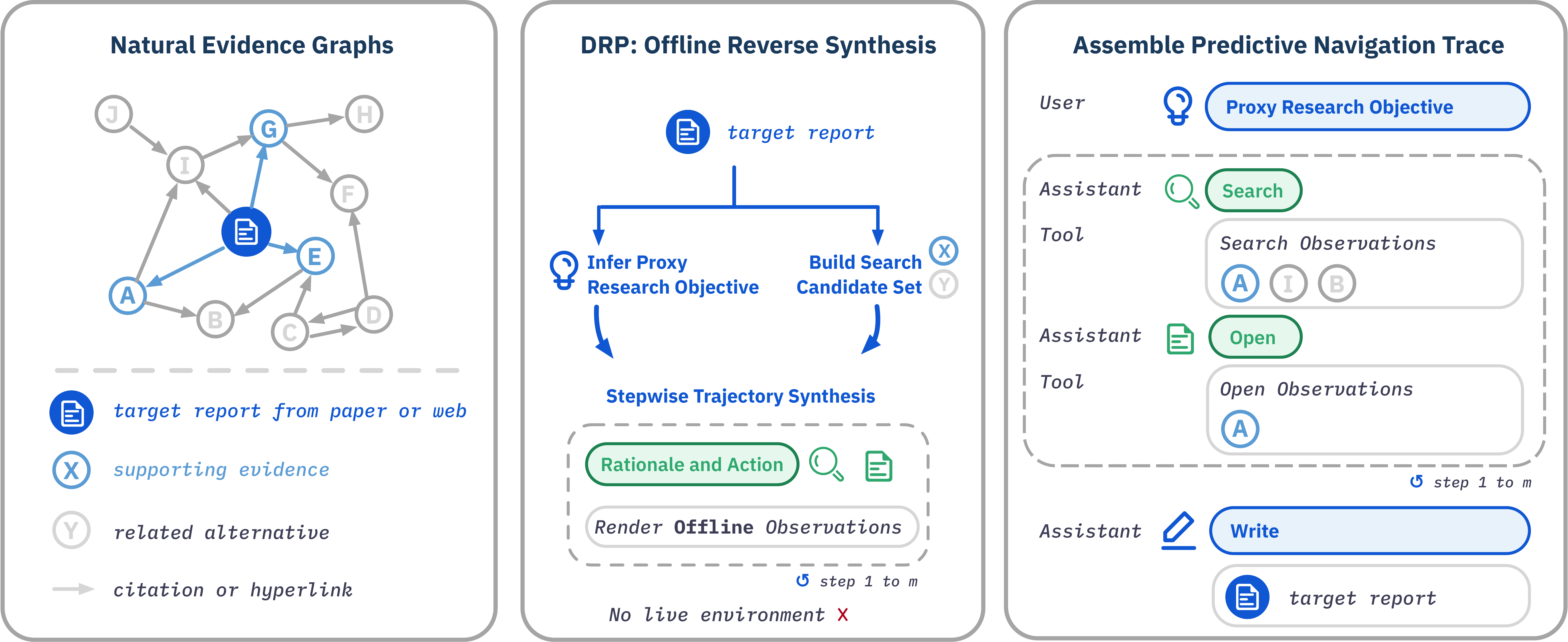}
\caption{DRP overview. Starting from a target report and its natural evidence graph, DRP infers a proxy research objective, builds search candidate sets from supporting evidence and related alternatives, and synthesizes a history-conditioned search--open--write trajectory. Search and open observations are rendered offline from graph neighborhoods and source documents, without a live retrieval environment or policy rollout.}
\label{fig:framework}
\end{figure*}

Language-model agents interleave reasoning with external actions and observations to solve multi-step tasks \citep{react}. Their capabilities are commonly elicited through post-training on task-specific, environment-grounded trajectories \citep{webgpt,drtulu}. Producing such data requires executable environments, long rollouts, repeated tool use, and outcome evaluation, making each usable trajectory costly. A representative setting in which these costs are amplified is deep research. A deep-research agent must reformulate queries, navigate multiple sources, select evidence, recover from unproductive exploration, and synthesize source-grounded reports. Recent methods train these abilities using teacher-generated trajectories or reinforcement learning in real or simulated search environments, often guided by report-level rubrics \citep{searchr1,deepresearcher,zerosearch,drtulu}. Although effective, they still depend on costly interaction, trajectory construction, or evaluation.

Yet document collections already contain evidence structures that arise independently of agent rollouts. Citation-bearing and hyperlinked passages associate synthesized text with documents that support it, while neighboring nodes in the citation or hyperlink graph provide related but unselected alternatives. These relations could potentially supervise the central decisions of deep research: what to search for, which documents to inspect, and how to synthesize the resulting evidence. Therefore, we ask: \emph{can naturally occurring relations among documents be transformed into rollout-free pretraining supervision, thereby improving the trajectory-sample efficiency of downstream agent training?} Our goal is not to replace downstream agent post-training, but to provide an initialization that reduces its dependence on expensive, environment-grounded trajectories.

We introduce \textbf{Deep Research Pretraining} (DRP), an offline framework that reverse-synthesizes such supervision from naturally occurring evidence structures rather than asking a teacher to execute a complete browsing rollout. Figure~\ref{fig:framework} summarizes the pipeline from natural evidence graphs to predictive-navigation trajectories. Starting from a citation-bearing or hyperlinked passage that serves as a target report, DRP constructs a proxy research objective, recovers linked sources, and draws graph-related alternatives from the surrounding citation or hyperlink topology. It then renders these ingredients into search--open--write trajectories. We call the learning objective \emph{predictive navigation}: conditioned on the research objective and the evidence observed so far, the model learns to predict what to search for, which documents to inspect, and how to synthesize the gathered evidence. Because candidate sets and document contents are assembled offline from the evidence graph and source corpus, constructing DRP data requires neither a live retrieval environment nor an executed policy rollout.

We instantiate DRP in two structurally distinct domains. \textbf{DRP-Paper} treats citation-bearing Related Work passages as target reports, infers proxy research objectives from their content, uses resolved cited papers as evidence, and samples topologically nearby but uncited papers as candidate alternatives. \textbf{DRP-Web} applies the same construction to Wikipedia passages: outgoing links whose pages substantively contribute to a passage form its evidence set after filtering incidental mentions, while neighboring nodes in the hyperlink graph provide related alternatives. Despite these domain-specific extraction rules, both variants render the resulting supervision with the same search--open--write action space and predictive-navigation objective. The graphs are used only during offline data construction; downstream agents operate with standard search and open tools.

We evaluate DRP as continual pretraining before agent SFT. Starting from Qwen3-14B-Base, we train separate models on 1B tokens of DRP-Paper or DRP-Web and then fine-tune them on controlled fractions of 13K GPT-5-generated deep-research trajectories from DR Tulu \citep{drtulu}. Here, sample efficiency is defined as performance as a function of the number of expert trajectories used for downstream training. Across five independently sampled subsets at both low-data budgets, DRP-Paper and DRP-Web consistently improve over matched no-DRP SFT on DeepResearch Bench. With one quarter of the trajectories, DRP-Web also exceeds a fixed full-data Base checkpoint. Positive low-data gains transfer to ResearchQA, WebWalkerQA, and SimpleQA, and the $1/16$-SFT effect persists on Qwen3-30B-A3B. Starting from matched $1/16$-SFT checkpoints, DRP-Web also remains ahead of Base throughout 80 steps of ARPO-style agentic RL. Source-matched raw-document and evidence-mismatch controls support the hypothesis that the gain comes from evidence-conditioned navigation rather than domain exposure or agent-format imitation alone.

Our contributions can be summarized as:
\begin{itemize}
    \item We introduce DRP, an offline reverse-synthesis framework that turns citation and hyperlink relations into history-conditioned search--open--write supervision without executing an agent rollout.
    \item We instantiate DRP on scholarly citation graphs and the Wikipedia hyperlink graph, and show that a fixed 1B-token DRP stage consistently improves downstream trajectory-SFT sample efficiency across independently sampled low-data subsets.
    \item Source-matched raw-CPT and evidence-mismatch controls support the hypothesis that DRP benefits from evidence-conditioned navigation rather than domain exposure or agent formatting alone.
\end{itemize}

\section{Method}
\label{sec:method}

\subsection{Problem Formulation}

Let $\mathcal{C}=\{d_v\}_{v\in\mathcal{V}}$ be a document corpus equipped with an evidence graph $\mathcal{G}=(\mathcal{V},\mathcal{E})$. A directed edge records a naturally occurring relation between two documents, such as a scholarly citation or a Wikipedia hyperlink. From an anchor document, we extract a citation-bearing or linked passage $r$ as a target report and recover a set of evidence vertices
\begin{equation}
E_r \subseteq \mathcal{V}
\end{equation}
whose associated documents $\{d_v:v\in E_r\}$ are referenced by and substantively related to that passage. We then infer a proxy research objective $u$ whose answer is expressed by $r$. Let $y=\mathrm{Normalize}(r,E_r)$ denote the final report target obtained from $r$ after retaining only resolvable evidence.

A deep-research trajectory contains both intermediate navigation decisions and a final report. We write a rendered trajectory as
\begin{equation}
\tau=\left(u,\left((z_i,a_i,o_i)\right)_{i=1}^{T},z_{\mathrm{w}},y\right),
\end{equation}
where $z_i$ is a reasoning trace, $a_i$ is a search or open action, $o_i$ is its observation, and $y$ is the final report preceded by a write rationale $z_{\mathrm{w}}$. At decision $i$, the model conditions only on the causal history
\begin{equation}
h_i=\left(u,\left((z_j,a_j,o_j)\right)_{j<i}\right)
\end{equation}
and predicts what to search for, which returned documents to open or ignore, and ultimately how to synthesize the acquired evidence. We call this learning problem \emph{predictive navigation}.

\subsection{Offline Reverse Synthesis}

Collecting $\tau$ in the usual forward direction requires executing a policy in a retrieval environment. DRP instead constructs it in reverse from $(r,E_r)$. We partition the evidence targets across search steps, $E_r=P_1\cup\cdots\cup P_m$. For step $i$, the constructor synthesizes a query $q_i$ and forms a search candidate set
\begin{equation}
K_i=P_i\cup N_i,
\end{equation}
where $P_i$ contains evidence vertices to acquire and $N_i$ contains vertices of topologically related documents that are plausible for the topic but are not selected for the target passage. The corresponding search observation renders titles and snippets for $K_i$. The open action selects $P_i$ and returns their source text; candidates in $N_i$ are observed but left unopened, providing an implicit selection contrast rather than explicit negative labels.
Section~\ref{sec:graph-instantiations} instantiates $E_r$ and the candidate alternatives $N_i$ using scholarly citation and Wikipedia hyperlink graphs.

Our default topology interleaves acquisition and inspection:
\begin{equation}
\texttt{S}(q_1)\rightarrow\texttt{O}(P_1)\rightarrow\cdots
\rightarrow\texttt{S}(q_m)\rightarrow\texttt{O}(P_m)
\rightarrow\texttt{W}(y).
\end{equation}
Consequently, each later query and its synthetic rationale can condition on documents opened earlier. We generate every rationale using only the information present in its rendered history $h_i$, although the offline constructor uses the hidden target report as an oracle for creating the supervision. We filter rationales that refer to the hidden construction process or unrevealed evidence. Let $\mathrm{Cite}(y)\subseteq\mathcal{V}$ denote the evidence vertices cited by $y$; normalization ensures that every cited evidence document has been opened:
\begin{equation}
\mathrm{Cite}(y)=\bigcup_{i=1}^{m}P_i=E_r.
\end{equation}
A batched control moves all open operations after the searches while preserving the underlying queries, evidence, and report; we analyze this topology separately.

This reverse construction requires no executed agent rollout. Queries are synthesized from the target contexts, candidate observations come from graph neighborhoods, and open observations are read directly from $\mathcal{C}$. Source links provide weak positive signals, graph neighborhoods provide candidate alternatives, and the source corpus provides offline observations. All three are obtained without calling a search engine or browser.

\subsection{Two Evidence-Graph Instantiations}
\label{sec:graph-instantiations}

\paragraph{DRP-Paper.}
We extract target reports from Related Work passages and resolve their citations to corpus documents through canonical scholarly identifiers. The resolved cited papers define $E_r$, while nearby but uncited papers in the citation graph supply the candidate alternatives $N_i$ at each search step. The report target retains only resolvable evidence and uses canonical paper citations.

\paragraph{DRP-Web.}
We extract target reports from Wikipedia passages and use outgoing links to identify candidate evidence pages. Because many hyperlinks are incidental mentions, we retain as $E_r$ only linked pages whose content is substantively used by the passage. Hyperlink-graph neighborhoods supply the candidate alternatives $N_i$. DRP-Paper and DRP-Web thus differ in how $E_r$ and $N_i$ are obtained, but share the same action space, interleaved construction, and learning objective.

\paragraph{Construction and corpus scale.}
DRP-Paper is built from the 2024 unarXive snapshot \citep{besrour-etal-2026-unarxive}, while DRP-Web uses the English FineWiki corpus \citep{finewiki}. We use Qwen3.6-35B-A3B-FP8 \citep{qwen36_35b_a3b} to infer proxy research objectives, generate search, open, and write rationales, and filter incidental Wikipedia links. Generation uses a maximum of 32,768 tokens, temperature 0.7, top-$p$ 0.8, and top-$k$ 20. Table~\ref{tab:drp-data} summarizes the resulting corpora.

\subsection{Masked Continual-Pretraining Objective}

Let $x_{1:L}$ be the tokenized trajectory and let $m_t$ indicate whether token $x_t$ is produced by the agent. DRP minimizes the masked causal language-model loss
\begin{equation}
\mathcal{L}_{\mathrm{DRP}}(\theta)
=-\frac{1}{\sum_{t=1}^{L}m_t}
\sum_{t=1}^{L}m_t\log p_{\theta}(x_t\mid x_{<t}).
\end{equation}
We set $m_t=1$ for assistant reasoning, tool calls and their arguments, and the final report, and $m_t=0$ for the system prompt, user objective, and search/open observations. The model therefore learns to predict evidence-conditioned decisions and synthesis rather than to reproduce the constructed environment responses.

\section{Experiments}
\label{sec:experiments}

\begin{figure*}[!ht]
\centering
\includegraphics[width=\textwidth]{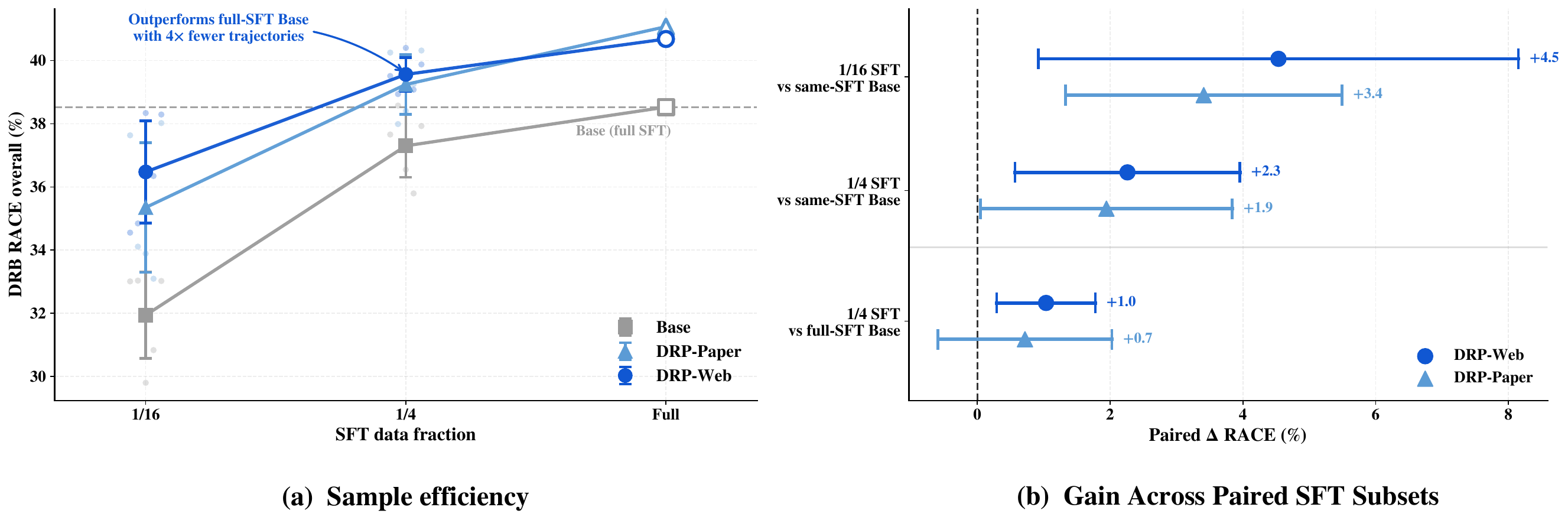}
\caption{DeepResearch Bench (English) sample efficiency with Qwen3-14B. \textbf{(a)} Mean scores over five independent SFT subsets; error bars are standard deviations, faint points are individual runs, and full-SFT points are fixed. \textbf{(b)} Paired score differences from Base at matched SFT fractions (top) and fixed full-SFT Base (bottom), with 95\% $t$-intervals over subsets.}
\label{fig:main}
\end{figure*}

\subsection{Experimental Setup}

\paragraph{Models and continual pretraining.}
Our primary model is Qwen3-14B-Base \citep{qwen3}, with Qwen3-30B-A3B-Base used for scale analysis. Unless stated otherwise, each DRP corpus uses a 1B-token CPT budget. We pack 128K-token sequences with a 2M-token global batch (500 steps) and train in BF16 with AdamW \citep{adamw} ($\beta_1=0.9$, $\beta_2=0.999$, $\epsilon=10^{-8}$), no weight decay, and gradient clipping at 1.0. The learning rate has 10\% warmup to $4\times10^{-5}$ followed by cosine decay to zero. A run takes approximately six hours on 128 H20 96GB GPUs. Canonical DRP-Web uses the interleaved search--open schedule and medium response budget analyzed in Section~\ref{sec:analysis}.

\paragraph{Downstream trajectory SFT.}
Following the DR Tulu SFT recipe \citep{drtulu}, we fine-tune each initialization for five epochs with 16K-token sequences, a 2M-token global batch, and the optimizer settings described above. We use three data budgets: $1/16$, $1/4$, and all 13,062 English GPT-5-generated trajectories from the public release.\footnote{\url{https://huggingface.co/datasets/rl-research/dr-tulu-sft-data}} These are environment-grounded expert demonstrations, so our claim concerns SFT sample efficiency rather than replacing agent training. At each low-data fraction, we sample five subsets (seeds 42, 2026, 5678, 31337, and 1234); all initializations receive the same trajectories within a seed. We report mean and standard deviation across runs. For contrasts involving subset-trained models, we compute the paired score difference $d_s$ for each seed and report $\bar d \pm t_{0.975,4}s_d/\sqrt{5}$, where $s_d$ is the sample standard deviation. This 95\% interval measures SFT-subset variation; full-data SFT is a fixed reference.

\paragraph{Downstream agentic RL.}
To test whether DRP remains useful beyond SFT, we initialize two RL runs from Base and DRP-Web checkpoints trained on the same fixed $1/16$ SFT subset. Both runs follow the ARPO deep-search setting \citep{arpo}, using the same 1K-example mixed hard-search training set and identical rollout, reward, and optimization configurations. We evaluate the post-SFT policies before RL and after 40 and 80 optimizer steps. The reported score is the unweighted mean of ResearchQA coverage, WebWalkerQA accuracy, and SimpleQA accuracy. Each run takes approximately 24 hours on 64 H20 GPUs.

\paragraph{Evaluation.}
Our main evaluation is the 50-question English subset of DeepResearch Bench (DRB) \citep{deepresearchbench}. We evaluate DRB with its official GPT-5.5 RACE evaluator.\footnote{\url{https://github.com/Ayanami0730/deep_research_bench}} Agents use the DR-Tulu protocol, Serper search, and Jina extraction, with at most 20 tool-call rounds and 32K output tokens. Decoding uses temperature 0.6, top-$p$ 0.95, and top-$k$ 20. We also report report-generation rate and mean search/open calls. Score differences are absolute points on this scale, not relative percentage changes. For single-checkpoint controls, we jointly resample question IDs, recompute the paired score difference, and take its 2.5th and 97.5th percentiles; unlike the $t$-intervals above, these paired-bootstrap intervals measure question uncertainty \citep{efron1994bootstrap}. For transfer, we evaluate fixed sets of 500 examples from ResearchQA \citep{researchqa}, WebWalkerQA \citep{webwalkerqa}, and SimpleQA \citep{simpleqa} with the same agent, reporting coverage for ResearchQA and accuracy otherwise.

\begin{table}[!h]
\centering
\small
\setlength{\tabcolsep}{2.5pt}
\begin{tabular}{@{}lrrr@{}}
\toprule
Corpus & Traj. & Available & Loss \% \\
\midrule
DRP-Paper & 78,557 & 1.99B & 8.67\% \\
DRP-Web & 8,997,455 & 172.46B & 7.73\% \\
\bottomrule
\end{tabular}
\caption{DRP corpus statistics before subsampling to the matched 1B-token CPT budget. Available is the total token count; Loss \% is the fraction included in the language-model loss.}
\label{tab:drp-data}
\end{table}

\subsection{Main Results}

\paragraph{DRP improves trajectory-SFT sample efficiency.}
Figure~\ref{fig:main} gives the central result. Across five independently sampled subsets at both $1/16$ and $1/4$ SFT, DRP-Paper and DRP-Web consistently improve over directly fine-tuning the base model. With only one quarter of the expert trajectories, DRP-Web exceeds the no-DRP full-data model by 1.0 point, with its 95\% interval above zero. DRP-Paper has a positive mean difference of 0.7 points, although its interval overlaps zero. The gain is largest in the lowest-data regime, where DRP also substantially reduces empty-report failures and produces a functioning tool-use policy. Figure~\ref{fig:main}(b) summarizes both matched-subset gains and direct comparisons with the fixed full-SFT Base model. The positive same-fraction gains show that the result recurs across independently sampled low-data training sets rather than depending on one selected subset.
Appendix~\ref{app:cost} estimates at least 51.8\% lower standardized data-construction cost.

\begin{figure*}[!t]
\centering
\includegraphics[width=\textwidth]{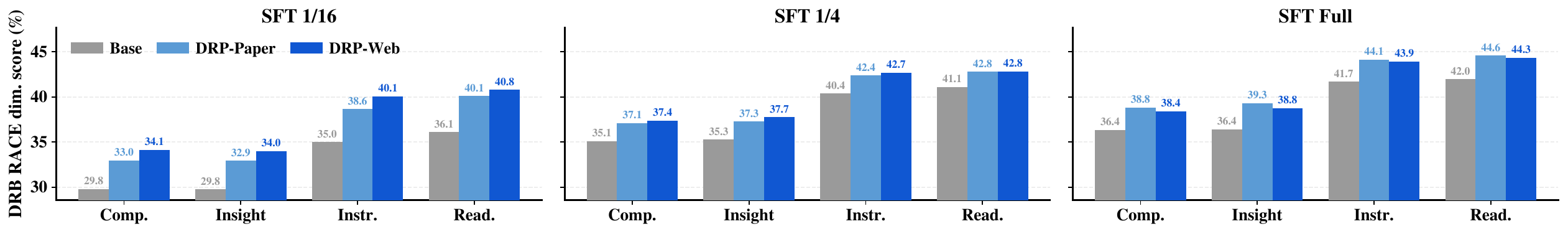}
\caption{RACE dimension scores. Low-data bars average five independently sampled, matched SFT subsets.}
\label{fig:race-dimensions}
\end{figure*}

\paragraph{The gain spans all RACE dimensions.}
Figure~\ref{fig:race-dimensions} decomposes the score into comprehensiveness, insight, instruction following, and readability. At both low-data fractions, DRP-Paper and DRP-Web improve every dimension over Base. The gains are distributed across content-oriented dimensions (comprehensiveness and insight) as well as instruction following and readability, rather than being concentrated in a single aspect of report generation.

\begin{figure*}[t]
\centering
\includegraphics[width=\textwidth]{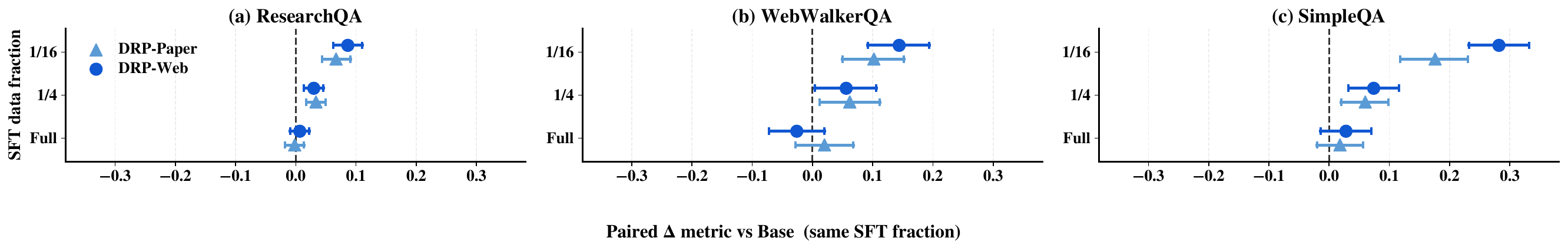}
\caption{Transfer over Base on 500 examples per benchmark. Markers show paired metric differences at matched SFT fractions; the metric is coverage for ResearchQA and accuracy otherwise. Bars are 95\% paired-bootstrap intervals over questions.}
\label{fig:transfer}
\end{figure*}

\paragraph{The gain transfers beyond long-form report evaluation.}
Figure~\ref{fig:transfer} shows that the benefit is not confined to long-form report evaluation. At both $1/16$ and $1/4$ SFT, DRP-Paper and DRP-Web have positive paired gains over Base on ResearchQA coverage, WebWalkerQA accuracy, and SimpleQA accuracy. Under full-data SFT, the effects become small and task dependent, with most confidence intervals overlapping zero. This attenuation is consistent with DRP acting as a sample-efficient initialization rather than directly optimizing each downstream benchmark.

\begin{figure}[t]
\centering
\includegraphics[width=0.5\columnwidth]{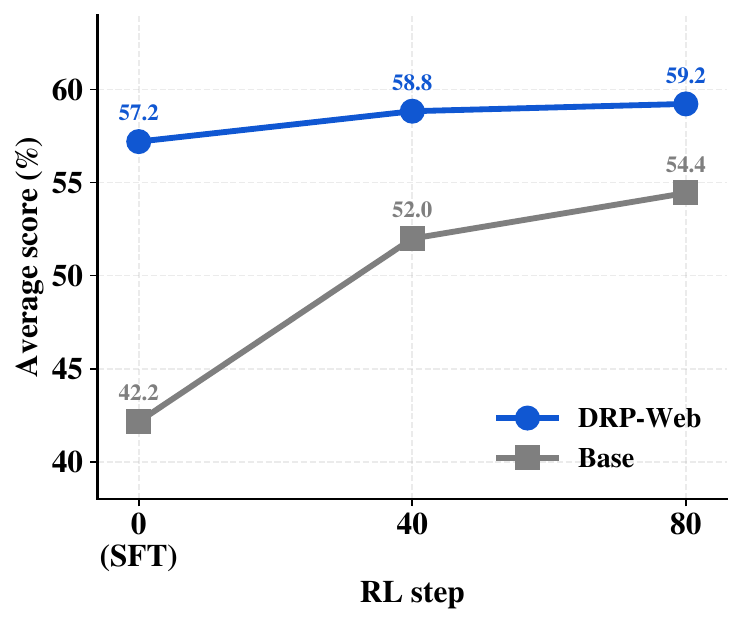}
\caption{Agentic RL from matched $1/16$-SFT checkpoints. Both initializations follow the same ARPO deep-search setting. Scores are the macro-average of ResearchQA coverage, WebWalkerQA accuracy, and SimpleQA accuracy.}
\label{fig:agentic-rl}
\end{figure}

\paragraph{The advantage persists through agentic RL.}
Figure~\ref{fig:agentic-rl} tests whether the stronger DRP initialization persists under subsequent policy optimization. Before RL, DRP-Web leads Base by 15.0 points (57.2 versus 42.2). The gap narrows but remains 6.8 points after 40 steps and 4.8 points after 80 steps. Base after 80 steps still does not reach the pre-RL DRP-Web checkpoint (54.4 versus 57.2), while DRP-Web itself improves to 59.2. Thus, DRP provides a stronger initial policy that remains advantageous under a fixed downstream RL budget and remains compatible with further policy optimization.

\section{Analysis}
\label{sec:analysis}

Unless stated otherwise, analysis experiments use Qwen3-14B, a 1B-token CPT budget, and the same fixed $1/16$ DR Tulu SFT subset; DRP-Web is the default corpus when no other corpus is named. All controls use the CPT and SFT optimization hyperparameters from Section~\ref{sec:experiments}, including sequence lengths, global batch sizes, learning-rate schedules, and SFT epochs. Thus, each comparison changes only the factor named by the ablation while holding the training budget and downstream supervision fixed.

\subsection{Raw Domain Pretraining Is Not Enough}

\begin{figure}[t]
\centering
\includegraphics[width=\columnwidth]{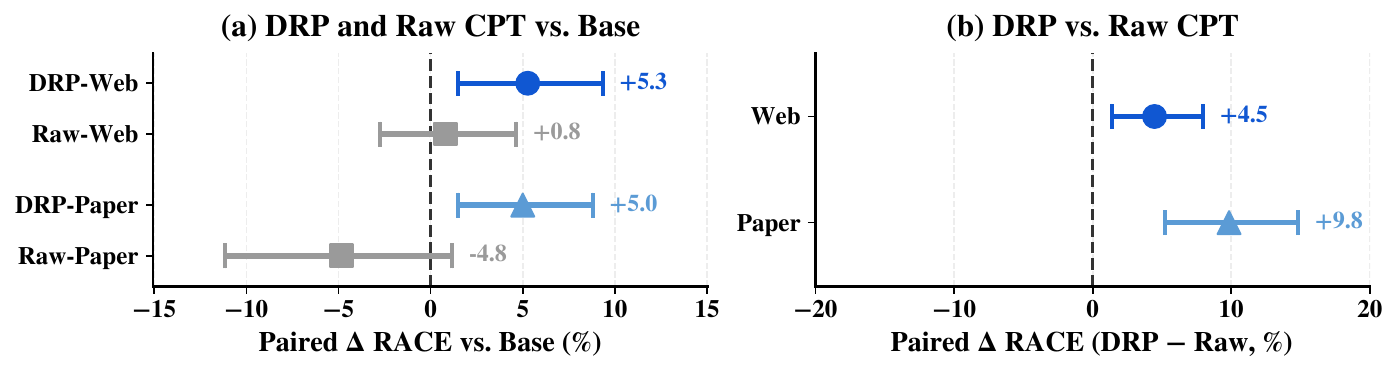}
\caption{Raw-domain CPT controls on DeepResearch Bench (English; Qwen3-14B; 1B-token CPT; $1/16$ SFT). \textbf{(a)} Paired score differences from Base. \textbf{(b)} DRP--Raw score differences with domain-matched source documents. Bars are 95\% paired-bootstrap intervals over questions.}
\label{fig:raw-cpt}
\end{figure}

We first test whether DRP's gain can be explained by exposure to the underlying source domains. Raw-Web and Raw-Paper train directly on the same source documents used to construct their respective DRP corpora, but contain no proxy queries, tool calls, or graph-derived relevance relations. Within each domain, we compare Base, raw-document CPT, and structured DRP under the same 1B-token CPT budget and downstream SFT setting. In Figure~\ref{fig:raw-cpt}(a), DRP-Web gains 5.3 points over Base, compared with 0.8 for Raw-Web; DRP-Paper gains 5.0 points, whereas Raw-Paper decreases by 4.8. The corresponding direct DRP--Raw advantages are 4.5 and 9.8 points, with both paired-bootstrap intervals above zero (Figure~\ref{fig:raw-cpt}(b)). Thus, source-matched domain exposure is not sufficient to account for the predictive-navigation gain.

\subsection{Evidence Alignment Matters}

\begin{figure}[t]
\centering
\includegraphics[width=\columnwidth]{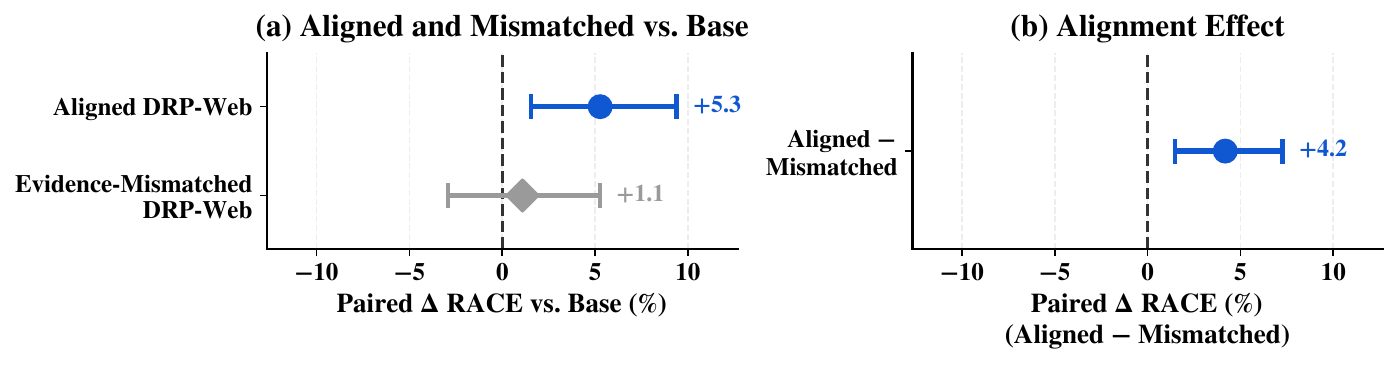}
\caption{Evidence-alignment control on DeepResearch Bench (English; Qwen3-14B; 1B-token DRP-Web CPT; fixed $1/16$-SFT subset). \textbf{(a)} Paired score differences from Base for aligned and evidence-mismatched DRP-Web. \textbf{(b)} Their direct score difference. Bars are 95\% paired-bootstrap intervals over questions.}
\label{fig:evidence-alignment}
\end{figure}

The raw-document controls establish an advantage for structured DRP over raw CPT, but do not isolate whether the constructed observations must be relevant to the model's decisions. We therefore perform a controlled ablation in our canonical DRP-Web analysis setting. Evidence-Mismatched DRP-Web preserves the proxy queries, synthetic reasoning, action schedule, tool-call syntax, response-length distribution, and training budget of aligned DRP-Web, but replaces graph-derived search and open observations with random in-domain evidence. Aligned DRP-Web gains 5.3 points over Base, while evidence-mismatched DRP-Web gains only 1.1; their direct difference is 4.2 points with its 95\% bootstrap interval above zero (Figure~\ref{fig:evidence-alignment}). Breaking query--observation--action alignment therefore removes most of the gain even though the agent-like surface form is preserved.

\subsection{Which Pseudo-operations Matter?}

\begin{table}[!t]
\centering
\small
\setlength{\tabcolsep}{2pt}
\begin{tabular}{@{}llrrr@{}}
\toprule
Component & Variant & Score & Reports & Calls \\
\midrule
Reference & No CPT & 33.0 & 86\% & 0.4 \\
\midrule
Operations & Search-only & 36.4 & 94\% & 3.5 \\
 & \texttt{SSSOW} & 37.3 & 100\% & 6.7 \\
 & \texttt{SOSOW}$^\dagger$ & \textbf{38.3} & 100\% & 5.7 \\
\midrule
Rendering & DR-Tulu$^\dagger$ & \textbf{38.3} & 100\% & 5.7 \\
 & Hermes & 34.0 & 86\% & 5.5 \\
\midrule
Loss target & Mask observations$^\dagger$ & \textbf{38.3} & 100\% & 5.7 \\
 & No mask & 30.2 & 96\% & 0.6 \\
\midrule
Response length & Long & 35.6 & 92\% & 3.9 \\
 & Mid$^\dagger$ & \textbf{38.3} & 100\% & 5.7 \\
 & Short & 36.4 & 96\% & 4.2 \\
\bottomrule
\end{tabular}
\caption{DRP-Web design ablations on DeepResearch Bench (English; Qwen3-14B; 1B-token CPT; fixed $1/16$-SFT subset). Reports is the valid-report rate, and Calls the mean search/open calls per question. \texttt{SSSOW} and \texttt{SOSOW} are mnemonic topology names rather than fixed-length traces. $^\dagger$ denotes canonical choices; bold marks each block's best score.}
\label{tab:design-ablation}
\end{table}

Let \texttt{S}, \texttt{O}, and \texttt{W} denote search, open, and write. The names \texttt{SSSOW} and \texttt{SOSOW} denote topology rather than a fixed operation count: the former contains a variable number of searches followed by open and write, while the latter contains a variable number of search--open rounds followed by write. Table~\ref{tab:design-ablation} compares these constructions with search-only traces. Search-only reaches 36.4, 3.4 points above no CPT (33.0). Batched \texttt{SSSOW} reaches 37.3, and interleaved \texttt{SOSOW} reaches 38.3 while using fewer downstream calls than the batched schedule (5.7 versus 6.7). This 1.0-point advantage is consistent with feedback-conditioned navigation, although the single fixed-subset comparison does not isolate every difference in the constructed histories.

\subsection{Protocol Rendering}

We render the same DRP records in either the compact DR-Tulu protocol used downstream or a multi-turn Hermes protocol that alternates assistant actions and tool observations. The underlying evidence, think traces, and tool calls are unchanged. Matching the downstream DR-Tulu protocol raises the score from 34.0 to 38.3 and report rate from 86\% to 100\% (Table~\ref{tab:design-ablation}). Protocol alignment therefore materially affects transfer efficiency. It is not sufficient to explain the full DRP gain, however, because evidence mismatch reduces the matched-format model's gain over Base from 5.3 to 1.1 points (Figure~\ref{fig:evidence-alignment}).

\subsection{Tool Response Length}

Tool observations are masked from the language-model loss. Shorter responses therefore reduce training tokens and increase the fraction of supervised tokens, but may omit evidence needed for grounded decisions. Under the fixed $1/16$ setting, the mid budget reaches a score of 38.3 with a 100\% report rate, compared with 36.4/96\% for short and 35.6/92\% for long (Table~\ref{tab:design-ablation}). Longer context therefore provides no monotonic benefit in this experiment. We use mid as the default practical tradeoff, but do not interpret this single-setting ordering as a universal optimum across downstream SFT budgets.

\subsection{Transfer Across Model Scales}

We repeat the $1/16$-SFT DRP-Paper and DRP-Web comparisons on Qwen3-30B-A3B-Base. Relative to matched no-CPT models, DRP-Paper and DRP-Web gain 5.0 and 5.3 points at 14B, and 4.7 and 3.1 points at 30B-A3B (Table~\ref{tab:model-scale}). The positive gains for both evidence graphs at both scales support transfer of the low-data effect; they do not establish a monotonic model-scaling law.

\begin{table}[!t]
\centering
\small
\setlength{\tabcolsep}{2.5pt}
\begin{tabular}{@{}llr@{}}
\toprule
Backbone & DRP corpus & $\Delta$ score (points) \\
\midrule
Qwen3-14B & Paper & +5.0 \\
 & Web & +5.3 \\
\midrule
Qwen3-30B-A3B & Paper & +4.7 \\
 & Web & +3.1 \\
\bottomrule
\end{tabular}
\caption{Model-scale transfer on DeepResearch Bench (English). Values are score gains from 1B-token DRP over no-CPT Base after matched SFT on one fixed $1/16$ subset; hence Qwen3-14B values differ from the five-subset means in Figure~\ref{fig:main}.}
\label{tab:model-scale}
\end{table}


\subsection{General Capability and Distribution Shift}

We test whether specializing CPT toward DRP harms general capabilities under a fixed 2B-token budget starting from Qwen3-14B-Base. The control uses 2B tokens of general midtraining data from the fully open OLMo 3 training flow \citep{olmo3}; the mixed condition replaces half of that budget with 1B DRP-Web tokens. Our 14-benchmark suite spans commonsense reasoning (e.g., ARC), mathematical reasoning (GSM8K), knowledge (MMLU-Redux), and reading comprehension (DROP); Appendix~\ref{app:general-capability} provides the complete suite, citations, and per-benchmark results. Table~\ref{tab:general-capability} shows that the aggregate score changes from 73.1 to 73.4. After matched $1/16$ DR-Tulu SFT, the mixed checkpoint improves the score from 33.1 to 36.8 (+3.7 points). Thus, replacing half of the general CPT budget with DRP-Web preserves aggregate general capability in this evaluation while improving downstream agent initialization.

\begin{table}[!t]
\centering
\small
\setlength{\tabcolsep}{4pt}
\begin{tabular}{@{}lrr@{}}
\toprule
CPT data & General avg. & DRB \\
\midrule
2B General & 73.1 & 33.1 \\
1B General + 1B DRP-Web & \textbf{73.4} & \textbf{36.8} \\
\bottomrule
\end{tabular}
\caption{General capability and downstream performance under a fixed 2B-token CPT budget. General average is measured after CPT; DRB is measured after matched $1/16$ SFT.}
\label{tab:general-capability}
\end{table}

\subsection{Masking Tool Observations Is Important}

Our default objective computes loss only on assistant reasoning, tool calls and arguments, and the final report; tool observations are context. A no-mask control additionally trains the model to predict the constructed search and open responses. Removing the mask lowers the score from 38.3 to 30.2, below the no-CPT score of 33.0, and reduces mean downstream tool calls from 5.7 to 0.6 (Table~\ref{tab:design-ablation}). This drop of 8.1 points shows that predicting constructed observations is detrimental in this setting; masking them focuses the objective on agent decisions and synthesis.

\section{Related Work}

\paragraph{Training and supervising research agents.}
WebGPT and ReAct establish browser-assisted question answering and the interleaving of language-model reasoning with external actions \citep{webgpt,react}. Recent deep-research systems train these behaviors through search-environment RL, teacher-generated trajectories, synthetic tasks, and report-level rubrics \citep{searchr1,deepresearcher,drtulu,quest}. ARPO further adapts policy optimization to multi-turn tool-use exploration \citep{arpo}. ZeroSearch and OpenResearcher reduce the cost of this supervision through simulated retrieval or a local search-and-browse environment \citep{zerosearch,openresearcher}. These approaches improve the downstream post-training process but still obtain navigation supervision from executed or simulated trajectories. DRP is complementary: it constructs an offline predictive-navigation initialization from existing evidence graphs before trajectory SFT or RL.

\paragraph{Continual and synthetic pretraining.}
Domain-adaptive pretraining shows that continued training on in-domain text can improve downstream adaptation \citep{dapt}; more recent work extends continual pretraining to agent behaviors or restructures source documents into higher-utility synthetic data \citep{agentfounder,finephrase}. WRAP++ uses hyperlink motifs to synthesize cross-document QA \citep{zhou2026wrapwebdiscoveryamplified}. These literature motivates both our pre-SFT intervention and our source-matched raw-CPT controls. DRP differs by converting document relations into history-conditioned evidence-acquisition decisions rather than only exposing the model to domain text or synthesizing cross-document knowledge.

\paragraph{Weak supervision from document structure.}
Naturally occurring document context has been used to generate pseudoqueries for weakly supervised retriever pretraining, while retrieved evidence structure can support dynamic planning at inference time \citep{ict,webweaver}. DRP connects these ideas at a different stage: citation and hyperlink relations provide offline supervision for an autoregressive policy spanning repeated search, evidence inspection, and report synthesis. Thus, the graph is neither only a single-step retrieval target nor an inference-time memory.

\section{Conclusion}

DRP turns naturally occurring citation and hyperlink relations into offline predictive-navigation supervision before downstream agent training. It renders graph-derived evidence and alternatives into history-conditioned search, open, and write decisions without live browser rollouts. Across five matched low-data subsets, both corpora improve trajectory sample efficiency; DRP-Web with one-quarter SFT surpasses the full-data no-DRP Base. This recipe uses 75\% fewer expert trajectories and reduces the standardized estimate of data-construction cost by at least 51.8\% (Appendix~\ref{app:cost}). Gains transfer to other benchmarks, persist through agentic RL, and span evidence graphs and model scales. Source-matched raw-CPT and evidence-mismatch controls attribute the benefit to navigation alignment rather than domain exposure or trace formatting, while general-data mixing preserves aggregate capability. DRP is therefore a complement, not a substitute, for expert-trajectory SFT and RL: it moves evidence acquisition into initialization so each downstream trajectory has greater value. This matters when environments are hard to deploy, feedback is delayed, or expert traces cover few workflows. Our evidence remains limited to English deep-research tasks, and the gains narrow with more supervision. Future work should test other evidence-bearing graphs and long-latency environments, and study how graph quality and alternative selection shape the benefits.


{\small
\bibliography{drp_aaai2027}
\bibliographystyle{iclr2026_conference}
}

\clearpage
\appendix
\setcounter{table}{0}
\setcounter{figure}{0}
\setcounter{equation}{0}
\setcounter{algorithm}{0}
\renewcommand{\thetable}{A\arabic{table}}
\renewcommand{\thefigure}{A\arabic{figure}}
\renewcommand{\theequation}{A\arabic{equation}}
\renewcommand{\thealgorithm}{A\arabic{algorithm}}
\providecommand{\theHtable}{}
\providecommand{\theHfigure}{}
\providecommand{\theHequation}{}
\providecommand{\theHalgorithm}{}
\renewcommand{\theHtable}{appendix.\arabic{table}}
\renewcommand{\theHfigure}{appendix.\arabic{figure}}
\renewcommand{\theHequation}{appendix.\arabic{equation}}
\renewcommand{\theHalgorithm}{appendix.\arabic{algorithm}}

\section{Reproducibility Details}
\label{app:reproducibility}

This appendix provides implementation details, statistical procedures, numerical
results, and precise definitions of the analysis controls.

\subsection{DRP Data Construction}

\paragraph{Source corpora.}
DRP-Paper is constructed from the 2024 unarXive snapshot
\citep{besrour-etal-2026-unarxive}. DRP-Web is constructed from the English
FineWiki corpus and its Wikipedia hyperlink structure \citep{finewiki}. After
filtering, DRP-Paper contains 78,557 trajectories and 1.99B available tokens;
DRP-Web contains 8,997,455 trajectories and 172.46B available tokens. Token
statistics use the Qwen3-8B tokenizer. All matched comparisons in the main paper
subsample each DRP corpus to a 1B-token continual-pretraining budget.

\paragraph{Candidate alternatives.}
For DRP-Paper, at each search round we construct the alternative pool from the
forward-citation neighbors of the evidence papers assigned to that round. We
exclude papers in the target evidence set and filter highly cited graph hubs,
which are often generic references (e.g., AdamW) rather than topic-specific
alternatives. For DRP-Web, we use the dual-link
($u\leftrightarrow v$) and co-mention
($u\rightarrow e\leftarrow v$ with $u\rightarrow v$) relations following
WRAP++ \citep{zhou2026wrapwebdiscoveryamplified}. Pages connected to the current
evidence through either motif form the alternative pool after excluding retained
evidence pages. For both corpora, alternatives are deduplicated over the
trajectory history: a document that has appeared in an earlier search
observation is not shown again as an alternative in a later round. Alternatives
appear in search observations but are not opened. Each search observation
contains 10 candidates: all evidence documents assigned to that round are
included, and alternatives fill the remaining slots.

\paragraph{Synthetic supervision.}
We use Qwen3.6-35B-A3B-FP8 \citep{qwen36_35b_a3b} to infer proxy research objectives, generate search,
open, and write rationales, and filter incidental Wikipedia links. Generation
uses a maximum of 32,768 tokens, temperature 0.7, top-$p$ 0.8, and top-$k$ 20.
The canonical medium response budget truncates rendered search observations to
400 tokens and open observations to 40,000 tokens. The response-length ablation
uses search/open limits of 200/20,000 tokens for the short condition and
800/80,000 tokens for the long condition.

\paragraph{Loss-bearing tokens.}
System prompts, user objectives, and synthetic search/open observations are
provided as context but masked from the language-model loss. Assistant reasoning,
tool calls and their arguments, and final reports receive loss. Under this
objective, loss-bearing tokens account for 8.67\% of DRP-Paper and 7.73\% of
DRP-Web.

Algorithm~\ref{alg:drp-construction} summarizes the canonical interleaved
construction. The evidence resolver and alternative builder are instantiated
with the citation-graph and hyperlink-graph rules described above.

\begin{algorithm}[t]
\caption{Offline construction of a DRP trajectory}
\label{alg:drp-construction}
\small
\textbf{Input}: Target passage $r$; graph $\mathcal{G}$; corpus $\mathcal{C}$;
corpus-specific resolver $\mathsf{Resolve}$ and alternative builder
$\mathsf{Alt}$; response budgets $B_{\mathrm{S}},B_{\mathrm{O}}$\\
\textbf{Output}: Predictive-navigation trajectory $\tau$ and token mask $m$
\begin{algorithmic}[1]
\STATE $E_r \leftarrow \mathsf{Resolve}(r,\mathcal{G},\mathcal{C})$
\STATE $y \leftarrow \mathsf{Normalize}(r,E_r)$; \quad
       $u \leftarrow \mathsf{InferObjective}(r)$
\STATE $(P_1,\ldots,P_M) \leftarrow \mathsf{Partition}(E_r,r)$
\STATE $\tau \leftarrow (u)$; \quad $h \leftarrow (u)$; \quad
       $H \leftarrow \emptyset$
\FOR{$i=1,\ldots,M$}
    \STATE $N_i \leftarrow
      \mathsf{Alt}(P_i,\mathcal{G})\setminus(E_r\cup H)$
    \STATE $N_i \leftarrow \mathsf{Select}(N_i,10-|P_i|)$
    \STATE $K_i \leftarrow P_i\cup N_i$
    \STATE $q_i \leftarrow \mathsf{SynthesizeQuery}(r,P_i,h)$
    \STATE $a_i^{\mathrm{S}}\leftarrow\mathsf{Search}(q_i)$; \quad
      $z_i^{\mathrm{S}}\leftarrow\mathsf{Rationale}(h,a_i^{\mathrm{S}})$
    \STATE $o_i^{\mathrm{S}}\leftarrow
      \mathsf{RenderSearch}(K_i,\mathcal{C};B_{\mathrm{S}})$
    \STATE Append $(z_i^{\mathrm{S}},a_i^{\mathrm{S}},o_i^{\mathrm{S}})$
      to $\tau$ and $h$; \quad $H\leftarrow H\cup K_i$
    \FOR{$v\in P_i$}
        \STATE $a^{\mathrm{O}}\leftarrow\mathsf{Open}(v)$; \quad
          $z^{\mathrm{O}}\leftarrow\mathsf{Rationale}(h,a^{\mathrm{O}})$
        \STATE $o^{\mathrm{O}}\leftarrow
          \mathsf{RenderOpen}(d_v;B_{\mathrm{O}})$
        \STATE Append $(z^{\mathrm{O}},a^{\mathrm{O}},o^{\mathrm{O}})$
          to $\tau$ and $h$
    \ENDFOR
\ENDFOR
\STATE Verify $\mathsf{Cite}(y)=\bigcup_{i=1}^{M}P_i$
\STATE $z_{\mathrm{w}}\leftarrow\mathsf{WriteRationale}(h)$; append
       $(z_{\mathrm{w}},\mathsf{Write},y)$ to $\tau$
\STATE Discard $\tau$ if a rationale reveals hidden construction state or
       unrevealed evidence
\STATE $m\leftarrow\mathsf{AgentTokenMask}(\tau)$
\RETURN $\tau,m$
\end{algorithmic}
\end{algorithm}


\subsection{DRP-Web Synthesis Prompts}
\label{app:synthesis-prompts}

The templates below document the LLM-generated components of canonical
interleaved DRP-Web; they do not replace the construction defined in the main
paper. Starting from target report $r$, the constructor filters its outgoing-link
pages to recover evidence set $E_r$, partitions that evidence into
$P_1,\ldots,P_m$, and augments each partition with graph-derived candidate
alternatives $N_i$ to form search candidate set $K_i=P_i\cup N_i$. Search and
open observations are then rendered offline from $K_i$ and the source corpus.
The prompts infer proxy research objective $u$, synthesize queries $q_i$, and
generate search, open, and write rationales for the interleaved
search--open--write trajectory. Runtime fields below use descriptive
angle-bracket placeholders aligned with this notation. Line breaks and dash
glyphs are normalized for typesetting without changing the instructions.

\paragraph{System prompts.}
Objective/query synthesis and write-rationale generation use:
{\small
\begin{verbatim}
You are a deep research agent simulating how a researcher explores an
encyclopedia.
Output exactly in the requested format.
\end{verbatim}
}
Search- and open-rationale generation use:
{\small
\begin{verbatim}
You are a deep research agent exploring an encyclopedia. You issue
search queries, read the pages you open, and reason about what to
look for next.
Output exactly in the requested format.
\end{verbatim}
}

\paragraph{Objective, evidence, and query synthesis.}
For DRP-Web, one call receives target report $r$ and the outgoing-link pages
considered while resolving $E_r$, represented by identifier, title, and lead
text. It filters incidental links, infers objective $u$, partitions retained
evidence into the per-round sets $P_i$, and synthesizes one query $q_i$ per
partition. These outgoing-link inputs are evidence-resolution candidates, not
the rendered search candidate sets $K_i$; graph-derived alternatives $N_i$ are
added separately.
{\small
\begin{verbatim}
You will write the report section below. It contains many hyperlinks,
but only SOME are genuine references whose content informs the
writing; others are incidental mentions (a person's birthplace, a
generic term) not actually discussed.

SECTION:
---
<target_report_r>
---

CANDIDATE linked pages (id + title + lead):
<outgoing_link_pages>

Step 1 -- SELECT only the pages whose content is substantively
used/discussed in the section (drop incidental/navigational links).

Step 2 -- Plan a realistic multi-round search (typically 2-6 rounds)
that would naturally surface the SELECTED pages.
- GROUP the selected pages BY SUB-TOPIC:
  each round issues ONE query that retrieves a COHERENT GROUP of
  related pages at once -- not one page at a time.
- A round may target a single page ONLY if it is genuinely its own
  sub-topic. Do NOT split a natural group into many one-page rounds.
- Write each QUERY at the SUB-TOPIC / category level, not as the
  proper name of one specific page.
- Every SELECTED page assigned to exactly one round; dropped pages
  appear nowhere.
- QUERY: 3-7 keywords, no questions, operators, or quotes.

Output EXACTLY (no markdown):
USER_TOPIC: <proxy_objective_u>
DROPPED: <comma-separated wiki:ID you judged incidental, or NONE>
ROUND 1
QUERY: <query_q_i>
GOLD: <comma-separated selected wiki:ID in evidence partition P_i>
ROUND 2
...
\end{verbatim}
}
Here, the prompt field \texttt{GOLD} denotes evidence partition $P_i$.

\paragraph{Search rationales.}
At search step $i$, the template generates rationale $z_i^{\mathrm{S}}$ from
objective $u$, query $q_i$, and the causal history $h_i$ available at that
position in the rendered trajectory. The query target $q_i$ is serialized after
$h_i$ and its rationale, so its prediction during CPT conditions on the prior
interaction history even though the constructor determines it offline. The
first-round template is:
{\small
\begin{verbatim}
You are a researcher beginning to investigate this topic:

USER_TOPIC: <proxy_objective_u>

The first aspect you want to explore is captured by this search:
QUERY: <query_q_1>

Write your opening first-person thinking: what the topic is
fundamentally about, how you plan to build up an understanding of it,
and why starting with this aspect makes sense.

Hard rules:
- Speak ONLY about the subject matter. Do NOT mention searching,
  queries, results, or sources -- only the topic and your plan.
- Do NOT restate the query verbatim. Do NOT mention any instructions.

Output EXACTLY:
THINK: <2-4 sentences>
\end{verbatim}
}
For subsequent rounds, the prompt includes the evidence opened in prior rounds:
{\small
\begin{verbatim}
You are a researcher building up a report on this topic:

USER_TOPIC: <proxy_objective_u>

Here is what you have LEARNED so far from your reading:
<rendered_history_h_i>

The aspect you want to explore NEXT is captured by this search:
QUERY: <query_q_i>

Write a SHORT first-person reflection that connects what you have
learned to this next step: what you now understand, and how exploring
this next aspect builds on or complements it. Frame it as forward
progress toward a fuller picture.

Hard rules:
- Speak ONLY about the subject matter and what you want to understand
  next.
- Do NOT mention searching, queries, results, pages, sources, or
  whether any earlier step "succeeded", "failed", "was
  relevant/irrelevant", or "lacked" anything. No commentary on the
  process -- only on the topic.
- Do NOT restate the query verbatim. Do NOT mention any instructions.

Output EXACTLY:
THINK: <2-4 sentences>
\end{verbatim}
}

\paragraph{Open rationales.}
The open-rationale template receives the rendered titles and snippets for
$K_i$, together with the titles of evidence documents in $P_i$ selected for
opening:
{\small
\begin{verbatim}
You are a researcher building up a report on this topic:

USER_TOPIC: <proxy_objective_u>

What you have LEARNED so far:
<rendered_history_h_i>

You just looked over these candidate pages
(title -- brief description):
<search_candidates_K_i>

You have decided to READ IN FULL the following pages:
<selected_evidence_titles_P_i>

Write a SHORT first-person reflection explaining WHY these particular
pages are the ones worth reading in full for this topic -- what each
contributes and how it advances your understanding. Distinguish them
from the others only implicitly (by focusing on their relevance), not
by naming which were skipped.

Hard rules:
- Speak ONLY about the subject matter and why these pages matter to
  it.
- Do NOT mention searching, queries, results, ranking, or whether
  anything "failed", "was irrelevant", or "lacked" anything. No
  process commentary.
- Do NOT mention that pages were provided or listed. Do NOT mention
  instructions.

Output EXACTLY:
THINK: <2-4 sentences>
\end{verbatim}
}

\paragraph{Write rationale.}
The final template generates $z_{\mathrm{w}}$ from the complete rendered
history before the model produces report target $y$:
{\small
\begin{verbatim}
You have finished gathering and reading sources for this topic:

USER_TOPIC: <proxy_objective_u>

You are about to write the report section.
In 2-3 sentences, think through how you will organize the writeup from
what you read -- what to lead with, how the pieces connect. Write
naturally as your own planning; do NOT mention instructions, lists,
or that any sources were provided.

Output EXACTLY:
THINK: <2-3 sentence writing plan>
\end{verbatim}
}

\paragraph{Rationale filtering.}
Each rationale may be regenerated up to three times. Consistent with the
filtering rule in the main method, we retain it only if it does not reveal the
hidden construction process or unrevealed evidence. Deterministic checks reject
references to supplied page lists or prompt instructions that would expose the
offline oracle.

\subsection{Continual Pretraining}

Table~\ref{tab:cpt-hparams} lists the continual-pretraining configuration.
Qwen3-14B-Base is the primary backbone, and Qwen3-30B-A3B-Base is used for the
model-scale analysis. We pack examples without crossing the 128K-token maximum
sequence length. Each 1B-token run contains 500 optimizer steps at a 2M-token
global batch size and takes approximately six hours on 128 H20 96GB GPUs.
CPT uses OLMo-core with training random seed 42.

\begin{table*}[t]
\centering
\small
\setlength{\tabcolsep}{6pt}
\begin{tabular}{@{}llll@{}}
\toprule
Hyperparameter & Value & Hyperparameter & Value \\
\midrule
Training tokens & 1B & Optimizer & AdamW \\
Maximum sequence length & 128K & Adam $(\beta_1,\beta_2)$ & $(0.9,0.999)$ \\
Global batch size & 2M tokens & Adam $\epsilon$ & $10^{-8}$ \\
Peak learning rate & $4\times10^{-5}$ & Weight decay & 0 \\
Learning-rate schedule & cosine decay to 0 & Gradient clipping & 1.0 \\
Warmup ratio & 0.1 & Precision & BF16 \\
\bottomrule
\end{tabular}
\caption{Continual-pretraining hyperparameters.}
\label{tab:cpt-hparams}
\end{table*}


\subsection{Downstream Agent SFT}

We use the 13,062 English GPT-5-generated trajectories released by DR Tulu
\citep{drtulu}.\footnote{\url{https://huggingface.co/datasets/rl-research/dr-tulu-sft-data}}
Following its SFT recipe, models are fine-tuned for five epochs on $1/16$,
$1/4$, or all trajectories (816, 3,265, and 13,062 examples, respectively)
using packed 16K-token sequences, a 2M-token global batch, BF16, and the
optimizer settings in Table~\ref{tab:cpt-hparams}. The optimization
configuration matches CPT except for the 16K maximum sequence length. SFT also
uses OLMo-core, with training random seed fixed at 42 for every run. Separately,
for each low-data fraction, we independently sample five subsets using data
seeds 42, 2026, 5678, 31337, and 1234. Within a data seed, every initialization
receives exactly the same downstream trajectories.

\subsection{Downstream Agentic RL}

We initialize matched RL runs from the Base and DRP-Web checkpoints trained on
the same fixed $1/16$ SFT subset. Both follow the ARPO deep-search setting
\citep{arpo} and are implemented with veRL using training random seed 42 and
the GRPO advantage estimator. We train for five epochs on the same 1K
hard-search examples with a global prompt batch of 64, a PPO mini-batch of 64,
and actor learning rate \(10^{-6}\). Prompts and responses are limited to 2,000
and 6,192 tokens, respectively. Synchronous vLLM tool rollouts use 12 samples
per prompt, six initial rollouts, beam size 2, branch probability 0.5, and
entropy weight 0.2. Both the KL-control and actor KL-loss coefficients are zero.
We use tensor parallelism of 2, dynamic batching, gradient checkpointing, and
validation before training; tool observations are excluded from the policy
loss. All RL data, rewards, rollouts, and optimization hyperparameters are
identical between initializations. We evaluate the post-SFT checkpoint and
checkpoints after 40 and 80 optimizer steps. Each run takes approximately
24 hours on 64 H20 GPUs.

\subsection{Agent Inference and Evaluation}

Agents use the DR-Tulu protocol with Serper for search and Jina for page
extraction. We permit at most 20 tool-call rounds and 32K output tokens. Decoding
uses temperature 0.6, top-$p$ 0.95, and top-$k$ 20.

The primary evaluation is the 50-question English subset of DeepResearch Bench
(DRB) \citep{deepresearchbench}. We evaluate DRB with its official GPT-5.5 RACE
evaluator.\footnote{\url{https://github.com/Ayanami0730/deep_research_bench}}
We report RACE overall
on its 0--100 scale as the primary score. Score differences are absolute points on this scale,
not relative percentage changes. A missing final report receives zero. Report rate is the percentage
of questions with a valid final report, and tool calls are the mean number of
search and open calls per question.

Transfer evaluations use fixed sets of 500 examples from ResearchQA
\citep{researchqa}, WebWalkerQA \citep{webwalkerqa}, and SimpleQA
\citep{simpleqa}. They use the same agent and a local Qwen3-30B judge. We report
coverage for ResearchQA and accuracy for WebWalkerQA and SimpleQA.


\subsection{ResearchQA Contamination Analysis}
\label{app:researchqa-contamination}

Because DRP-Paper is derived from scholarly text, we test all 776 ResearchQA
queries for verbatim overlap with the delivered DRP-Paper corpus, comprising
78.6K trajectories from 55,648 distinct arXiv source papers. For each query, we
check whether any of its word $k$-grams appears in the DRP-Paper training text.

\begin{table}[t]
\centering
\small
\setlength{\tabcolsep}{6pt}
\begin{tabular}{@{}lrr@{}}
\toprule
$k$-gram & Queries with overlap & Max. containment \\
\midrule
8-gram & 0/776 & 0.000 \\
13-gram & 0/776 & 0.000 \\
\bottomrule
\end{tabular}
\caption{Verbatim overlap between ResearchQA queries and DRP-Paper training
text.}
\label{tab:researchqa-contamination}
\end{table}

No ResearchQA query shares even one 8- or 13-word span with the DRP-Paper
training corpus, providing no evidence of verbatim benchmark-query leakage.

\section{Statistical Procedures}
\label{app:statistics}

\subsection{Variation Across Sampled SFT Subsets}

The five low-data runs at each fraction vary the sampled SFT subset rather than
the training random seed. Let $M_{a,s}$ be the score of initialization $a$ after
SFT on the subset sampled with seed $s$, and let $b$ denote the no-CPT Base
initialization. For a matched comparison, we first compute
\begin{equation}
d_s=M_{a,s}-M_{b,s}.
\end{equation}
We report the mean paired score gain $\bar d$ and its 95\% Student-$t$ confidence
interval,
\begin{equation}
\bar d \ \mathbin{\pm}\ t_{0.975,4}\frac{s_d}{\sqrt{5}},
\qquad t_{0.975,4}=2.776,
\end{equation}
where $s_d$ is the sample standard deviation of the five paired differences.
This interval measures how the estimated gain changes when the downstream SFT
subset changes. It does not measure training-seed variation or question-level
evaluation uncertainty.

For the comparison between one-quarter-SFT DRP and full-SFT Base, the Base score
is a fixed reference. We compute $d_s=M_{a,s}^{1/4}-M_b^{\mathrm{full}}$ for each
of the five sampled one-quarter subsets and apply the same $t$ interval. Thus,
the interval measures subset variation around the fixed full-data reference.

\subsection{Question-Level Paired Bootstrap}

For single-checkpoint analyses, both compared systems answer the same DRB
questions. We jointly resample question IDs with replacement, preserving the
pairing between systems, recompute the mean score difference for each bootstrap
replicate, and report the 2.5th and 97.5th percentiles. This paired bootstrap
captures uncertainty over the evaluated questions while holding the trained
checkpoints fixed. It is therefore complementary to, rather than interchangeable
with, the subset-level $t$ intervals above. We use 10,000 bootstrap replicates
with random seed 42.

\subsection{Repeated-Evaluation Stability}

To quantify variation from stochastic agent rollouts and evaluation, we
independently evaluate the same DRP-Web full-SFT checkpoint three times under
the same evaluation configuration. Table~\ref{tab:repeated-evaluation} shows a
mean score of 40.64, a sample standard deviation of 0.46, and a range of 0.91
points. This evaluation-only spread is smaller than the matched low-data gains
in the main results, but it is measured for one full-SFT checkpoint and does
not replace the paired uncertainty analyses above.

\begin{table}[t]
\centering
\small
\setlength{\tabcolsep}{3pt}
\begin{tabular}{@{}lclrr@{}}
\toprule
SFT fraction & Runs & DRB scores & Mean & Range \\
\midrule
Full & 3 & 41.09 / 40.18 / 40.65 & 40.64 & 0.91 \\
\bottomrule
\end{tabular}
\caption{Repeated evaluations of the same DRP-Web full-SFT checkpoint. Scores
use the 0--100 scale.}
\label{tab:repeated-evaluation}
\end{table}


\section{Additional Main Results}
\label{app:additional-main}

\subsection{Per-Subset DeepResearch Bench Results}

Tables~\ref{tab:main-per-subset} and~\ref{tab:main-contrasts} give the
numerical results underlying the main-paper sample-efficiency figure. The five
columns at each low-data fraction correspond to independently sampled, matched
SFT subsets, not different training seeds. All values are absolute score points
on the 0--100 scale.

\begin{table*}[t]
\centering
\small
\setlength{\tabcolsep}{5pt}
\begin{tabular}{@{}llrrrrrr@{}}
\toprule
Initialization & SFT & Seed 42 & Seed 2026 & Seed 5678 & Seed 31337 &
Seed 1234 & Mean $\pm$ SD \\
\midrule
Base & $1/16$ & 33.0 & 33.0 & 29.8 & 30.8 & 33.0 & $31.94 \pm 1.37$ \\
 & $1/4$ & 37.7 & 38.6 & 36.5 & 35.8 & 37.9 & $37.30 \pm 1.00$ \\
 & Full & -- & -- & -- & -- & -- & 38.53 \\
\midrule
DRP-Paper & $1/16$ & 37.6 & 34.1 & 33.9 & 33.1 & 38.0 &
$35.35 \pm 2.06$ \\
 & $1/4$ & 40.3 & 38.0 & 38.4 & 39.3 & 40.3 & $39.25 \pm 0.94$ \\
 & Full & -- & -- & -- & -- & -- & 41.08 \\
\midrule
DRP-Web & $1/16$ & 34.6 & 34.8 & 38.3 & 36.3 & 38.3 &
$36.48 \pm 1.62$ \\
 & $1/4$ & 39.5 & 38.9 & 40.4 & 39.1 & 39.9 & $39.56 \pm 0.53$ \\
 & Full & -- & -- & -- & -- & -- & 40.69 \\
\bottomrule
\end{tabular}
\caption{Per-subset DeepResearch Bench results for the main Qwen3-14B
experiment. Low-data rows report five independently sampled SFT subsets and
their mean and standard deviation. Full-SFT values are fixed single-checkpoint
references.}
\label{tab:main-per-subset}
\end{table*}

\begin{table}[t]
\centering
\small
\setlength{\tabcolsep}{3pt}
\begin{tabular}{@{}llrr@{}}
\toprule
Initialization & Comparison & Mean $\Delta$ & 95\% CI \\
\midrule
DRP-Paper & $1/16$ vs.\ matched Base & $+3.41$ & $[+1.33,+5.49]$ \\
 & $1/4$ vs.\ matched Base & $+1.94$ & $[+0.05,+3.84]$ \\
 & $1/4$ vs.\ full-SFT Base & $+0.72$ & $[-0.59,+2.03]$ \\
\midrule
DRP-Web & $1/16$ vs.\ matched Base & $+4.54$ & $[+0.92,+8.15]$ \\
 & $1/4$ vs.\ matched Base & $+2.26$ & $[+0.57,+3.96]$ \\
 & $1/4$ vs.\ full-SFT Base & $+1.04$ & $[+0.29,+1.78]$ \\
\bottomrule
\end{tabular}
\caption{Subset-level paired score contrasts for the main experiment.
Intervals are 95\% Student-$t$ intervals over the five paired SFT subsets, as
defined in Section~\ref{app:statistics}. Full-SFT Base is a fixed reference.}
\label{tab:main-contrasts}
\end{table}

\subsection{Full-SFT Reference}

The full-SFT results come from single trained checkpoints and are therefore
shown as fixed training references rather than as estimates over the five
low-data subsets. The main claim concerning four-times fewer trajectories is
supported by comparing each one-quarter-SFT DRP run with the same full-SFT Base
checkpoint and its designated evaluation. We keep this distinction explicit to
avoid interpreting the displayed interval as training-seed uncertainty for full
SFT. Repeated evaluation of the DRP-Web full-SFT checkpoint is reported
separately in Table~\ref{tab:repeated-evaluation}.

\subsection{RACE Dimension Breakdown}

Table~\ref{tab:race-dimensions} gives the numerical values underlying the RACE
dimension figure in the main paper. Both DRP variants improve comprehensiveness,
insight, instruction following, and readability at $1/16$ and $1/4$ SFT. This
rules out an explanation in which the overall gain is caused solely by style or
report validity.

\begin{table*}[t]
\centering
\small
\setlength{\tabcolsep}{7pt}
\begin{tabular}{@{}llrrrrr@{}}
\toprule
Initialization & SFT & Comp. & Insight & Instr. & Read. & Overall \\
\midrule
Base & $1/16$ & 29.8 & 29.8 & 35.0 & 36.1 & 31.9 \\
 & $1/4$ & 35.1 & 35.3 & 40.4 & 41.1 & 37.3 \\
 & Full & 36.4 & 36.4 & 41.7 & 42.0 & 38.5 \\
\midrule
DRP-Paper & $1/16$ & 33.0 & 32.9 & 38.6 & 40.1 & 35.3 \\
 & $1/4$ & 37.1 & 37.3 & 42.4 & 42.8 & 39.2 \\
 & Full & 38.8 & 39.3 & 44.1 & 44.6 & 41.1 \\
\midrule
DRP-Web & $1/16$ & 34.1 & 34.0 & 40.1 & 40.8 & 36.5 \\
 & $1/4$ & 37.4 & 37.7 & 42.7 & 42.8 & 39.6 \\
 & Full & 38.4 & 38.8 & 43.9 & 44.3 & 40.7 \\
\bottomrule
\end{tabular}
\caption{Numerical DRB RACE-dimension results. Values at $1/16$ and $1/4$
SFT are means over five independently sampled, matched SFT subsets; full-SFT
values are fixed single-run references. All scores use the 0--100 scale.}
\label{tab:race-dimensions}
\end{table*}

\subsection{General-Capability Scores}
\label{app:general-capability}

\begin{figure*}[t]
\centering
\includegraphics[width=\textwidth]{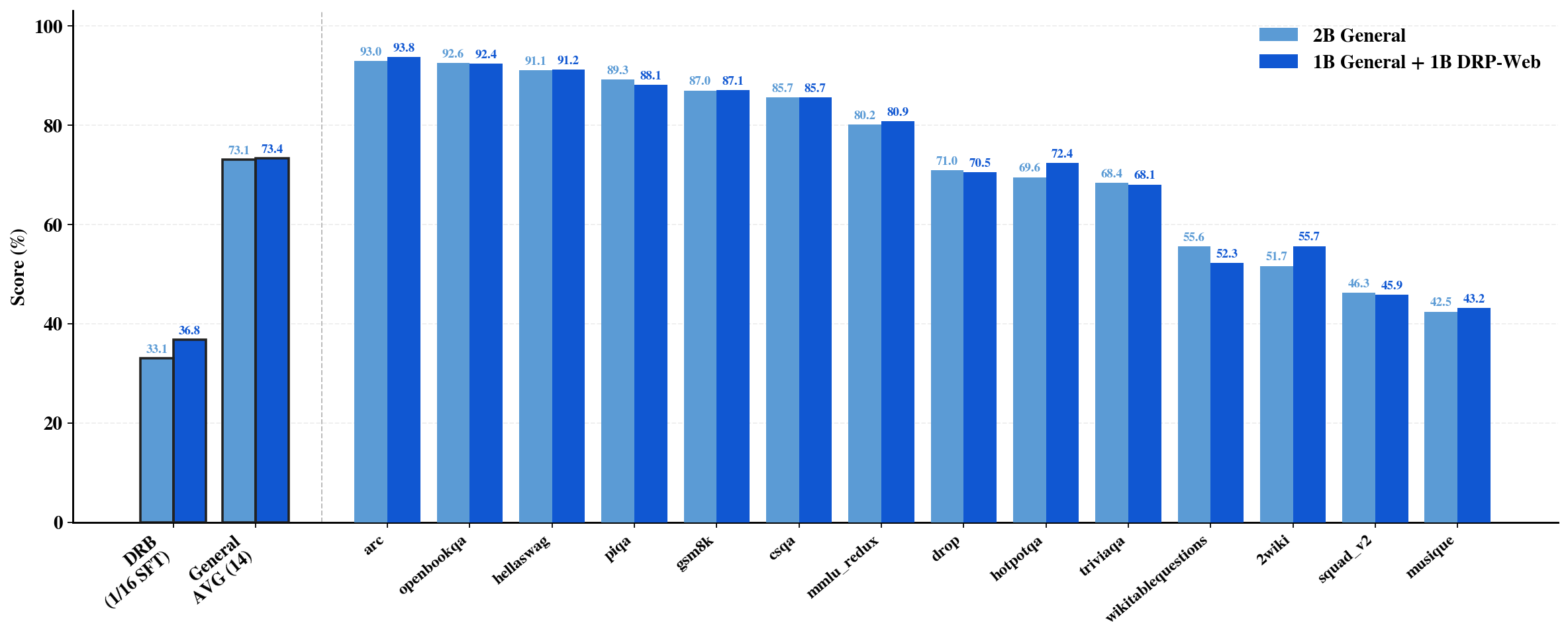}
\caption{Per-benchmark general-capability results and downstream
DeepResearch Bench performance under a fixed 2B-token Qwen3-14B CPT budget.
The control uses 2B general midtraining tokens; the mixed model uses 1B general
and 1B DRP-Web tokens. General benchmarks are evaluated after CPT, and
DeepResearch Bench after matched $1/16$ DR-Tulu SFT.}
\label{fig:general-capability-full}
\end{figure*}

The general-capability suite comprises ARC, OpenBookQA, HellaSwag, PIQA,
GSM8K, CommonsenseQA, MMLU-Redux, DROP, HotpotQA, TriviaQA,
WikiTableQuestions, 2WikiMultiHopQA, SQuAD 2.0, and MuSiQue
\citep{arc,openbookqa,hellaswag,piqa,gsm8k,commonsenseqa,mmluredux,drop,
hotpotqa,triviaqa,wikitq,twowiki,squad2,musique}. Figure~\ref{fig:general-capability-full}
visualizes the per-benchmark results, and Table~\ref{tab:general-capability-full}
gives their numerical values. Replacing half of the 2B-token general CPT
budget with DRP-Web changes the 14-benchmark average from 73.1 to 73.4, while
improving DeepResearch Bench after matched $1/16$ trajectory SFT from 33.1 to
36.8.

\begin{table*}[t]
\centering
\small
\setlength{\tabcolsep}{5pt}
\begin{tabular}{@{}lrrlrr@{}}
\toprule
Benchmark & General & Mixed & Benchmark & General & Mixed \\
\midrule
ARC & 93.0 & \textbf{93.8} & DROP & \textbf{71.0} & 70.5 \\
OBQA & \textbf{92.6} & 92.4 & HotpotQA & 69.6 & \textbf{72.4} \\
HellaSwag & 91.1 & \textbf{91.2} & TriviaQA & \textbf{68.4} & 68.1 \\
PIQA & \textbf{89.3} & 88.1 & WikiTableQ & \textbf{55.6} & 52.3 \\
GSM8K & 87.0 & \textbf{87.1} & 2Wiki & 51.7 & \textbf{55.7} \\
CSQA & \textbf{85.7} & \textbf{85.7} & SQuAD v2 & \textbf{46.3} & 45.9 \\
MMLU-R & 80.2 & \textbf{80.9} & MuSiQue & 42.5 & \textbf{43.2} \\
\midrule
General avg. & 73.1 & \textbf{73.4}
 & DRB ($1/16$ SFT) & 33.1 & \textbf{36.8} \\
\bottomrule
\end{tabular}
\caption{General-capability preservation and downstream utility under a fixed
2B-token Qwen3-14B CPT budget. General uses 2B general tokens; Mixed uses 1B
general plus 1B DRP-Web tokens. General benchmarks evaluate post-CPT
checkpoints; DRB evaluates both after matched $1/16$ DR-Tulu SFT. Bold marks
the higher score for each benchmark; both entries are bold for ties.}
\label{tab:general-capability-full}
\end{table*}

\FloatBarrier

\section{Synthetic-Data Cost Accounting}
\label{app:cost}

We estimate the cost of constructing training data, separately from the cost of
training on that data. The comparison uses public list prices as a common
reference rather than our historical expenditure: \$10 per million output
tokens for GPT-5\footnote{\url{https://developers.openai.com/api/docs/models/gpt-5}}
and \$1 per million output tokens for Qwen3.6-35B-A3B on
OpenRouter.\footnote{\url{https://openrouter.ai/qwen/qwen3.6-35b-a3b}}
All token volumes are serialized-data estimates computed as characters divided
by 4.5.

For a conservative, reproducible comparison, we price only visible output text.
Input-token billing and external search/retrieval charges are excluded because
historical request logs are unavailable. This choice favors expert SFT: an
online GPT-5 trajectory repeatedly consumes the growing interaction history as
input and invokes real search/open services, whereas DRP observations are
rendered directly from the evidence graph and source corpus. We nevertheless
count \emph{all} 80.42M non-observation tokens in the DRP-Web 1B slice at the
Qwen output price. This is an upper bound because the category includes
source-derived target reports in addition to generated proxy objectives,
rationales, and tool arguments.

\begin{table}[t]
\centering
\small
\setlength{\tabcolsep}{4pt}
\begin{tabular}{@{}lrrr@{}}
\toprule
Recipe & Score & Expert traj. & Data cost \\
\midrule
Base + Full SFT & 38.5 & 13,062 & \$347.6 \\
DRP-Web + $1/4$ SFT & \textbf{39.6} & 3,265 & $\leq$\$167.5 \\
\bottomrule
\end{tabular}
\caption{Standardized data-construction cost. Full SFT contains an estimated
34.76M visible GPT-5 output tokens. The DRP recipe combines 8.71M GPT-5 output
tokens (\$87.1) with an upper bound of 80.42M Qwen output-equivalent tokens
(at most \$80.4). Thus, it reduces the output-price proxy by at least 51.8\%
while using 75\% fewer expert trajectories and obtaining a higher
DeepResearch Bench score. Training costs are outside this accounting boundary.}
\label{tab:cost-recipes}
\end{table}

\begin{figure}[t]
\centering
\includegraphics[width=0.5\linewidth]{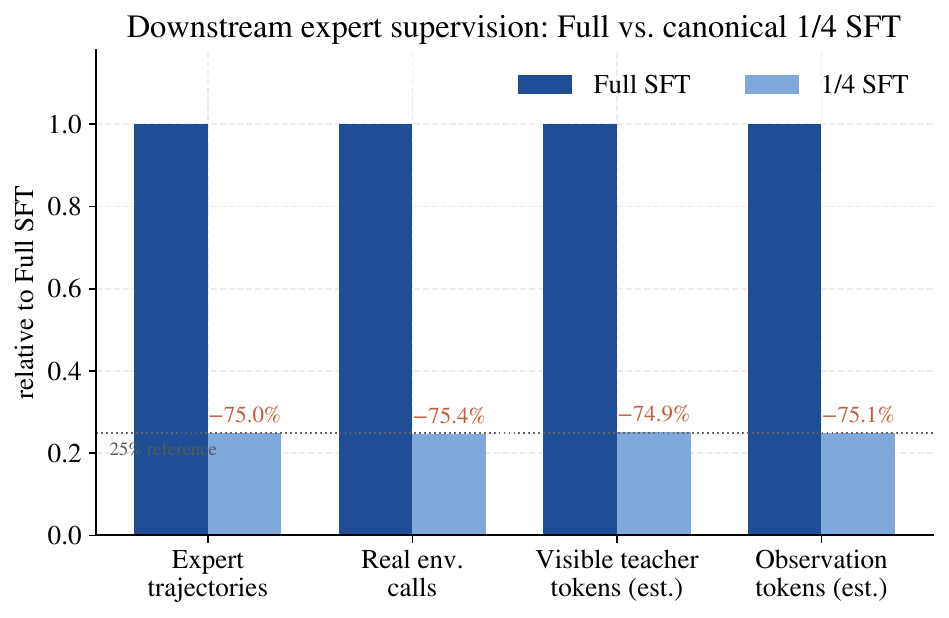}
\caption{The canonical one-quarter SFT subset uses approximately 75\% fewer
expert trajectories, real environment calls, visible teacher-output tokens,
and observation tokens than full SFT. DRP-Web constructs its search/open
observations offline and therefore adds no real environment calls.}
\label{fig:cost-supervision}
\end{figure}

The resulting comparison isolates the resource that DRP is designed to save:
expensive downstream expert supervision. Base + Full SFT costs approximately
\$347.6 under the output-price proxy. DRP-Web synthesis plus one-quarter SFT
costs at most \$167.5, a reduction of at least 51.8\%, while improving the score
from 38.5 to 39.6. It also reduces real environment calls from 47,147 to 11,597
(75.4\%). The advantage is therefore not that pretraining is free, but that an
inexpensive offline generator and naturally occurring evidence graphs can
amortize agent supervision before downstream post-training. This is especially
relevant when task-specific expert trajectories are scarce or when online
environment feedback is slow.

\FloatBarrier

\section{Construction and Ablation Details}
\label{app:ablations}

\subsection{Raw-Domain Controls}

Raw-Web and Raw-Paper use the same source-document pools as DRP-Web and
DRP-Paper, respectively. They match the 1B-token CPT budget but remove proxy
queries, reasoning traces, tool calls, and graph-derived relevance relations.
Consequently, the DRP--Raw comparison isolates the effect of restructuring the
same domains into predictive-navigation supervision.

\subsection{Evidence-Mismatched DRP-Web}

Evidence-Mismatched DRP-Web preserves the proxy objectives, generated reasoning,
action schedule, protocol rendering, response-length distribution, and training
budget of canonical DRP-Web. We replace the graph-derived search and open
observations with random in-domain evidence. The ablation therefore breaks the
alignment among query, observation, and subsequent action while preserving the
agent-like surface form.

\subsection{Pseudo-operation Topologies}

We use \texttt{S}, \texttt{O}, and \texttt{W} for search, open, and write. The
search-only condition omits full-document open observations. The batched
\texttt{SSSOW} and interleaved \texttt{SOSOW} names denote topology rather than
fixed-length traces. The former performs a variable number of searches before
opening selected evidence and writing; the latter performs a variable number of
search--open rounds before writing. Interleaving allows later synthetic
rationales and queries to condition on previously opened evidence.

\subsection{Protocol Rendering}

The DR-Tulu and Hermes conditions contain the same objectives, think traces,
tool calls, observations, and final reports. They differ only in serialization.
DR-Tulu groups the constructed interaction in the compact downstream protocol,
whereas Hermes alternates assistant tool-call turns and tool-response turns.
This control tests protocol alignment rather than a change in supervision.

\subsection{Response Length and Loss Masking}

The response-length ablation changes the amount of evidence rendered in masked
tool observations. Shorter responses reduce training cost and increase the
fraction of loss-bearing tokens, but may omit information required for the next
decision or final report. Separately, the no-mask condition computes loss on
constructed tool observations in addition to assistant outputs; all other data
and optimization settings remain fixed.

\section{Additional Limitations}
\label{app:limitations}

The current experiments use English downstream trajectories and one released
trajectory source. The primary DRB evaluation has 50 English questions, making
paired evaluation and transparent uncertainty reporting especially important.
The five-run design measures downstream subset sensitivity but does not estimate
the full interaction among subset sampling, training randomness, decoding
randomness, and judge variation. Most construction ablations use one fixed SFT
subset and paired question-level bootstrap intervals. Finally, DRP's gain
attenuates with more downstream SFT data.

\end{document}